\documentclass{article}

\usepackage{amsmath}
\usepackage{graphicx}
\usepackage[hidelinks]{hyperref}
\usepackage{float}

\title{Fair inference on error-prone outcomes}
\author{Laura Boeschoten\footnote{Shared first authorship. These authors contributed equally to the current work. Correspondence regarding this article should be sent to \texttt{l.boeschoten@uu.nl} and \texttt{e.vankesteren1@uu.nl.}} \and Erik-Jan van Kesteren\footnotemark[1] \and Ayoub Bagheri \and Daniel L. Oberski\\}

\date{Utrecht University \\ Department of Methodology \& Statistics}

\begin{document}
\maketitle
\begin{abstract}
Fair inference in supervised learning is an important and active area of research,  yielding a range of useful methods to assess and account for fairness criteria when predicting ground truth targets. As shown in recent work, however, when target labels are error-prone, potential prediction unfairness can arise from measurement error.  In this paper, we show that, when an error-prone proxy target is used, existing methods to assess and calibrate fairness criteria do not extend to the true target variable of interest. To remedy this problem, we suggest a framework resulting from the combination of two existing literatures: fair ML methods, such as those found in the counterfactual fairness literature on the one hand, and, on the other, measurement models found in the statistical literature. We discuss these approaches and their connection resulting in our framework. In a healthcare decision problem, we find that using a latent variable model to account for measurement error removes the unfairness detected previously. \break
\end{abstract}

\section{Introduction}
Supervised learning is used to guide human decisions across a wide range of different fields. In sensitive areas such as healthcare or criminal justice, a key issue is that decisions based on such decisions are equitable and fair. To this end, an active area of research investigates how  fairness criteria can be incorporated into supervised learning  \cite{berk2018fairness,corbett2017algorithmic,dwork2012fairness,kleinberg2016inherent,kusner_counterfactual_2017,verma_fairness_2018}. This literature has focused on supervised learning for a single objective, assumed to be the target variable of interest. Recently, however, \cite{obermeyer_dissecting_2019} observed that, even when substantial care has been taken to develop a prediction algorithm, unfairness in the predictions can still result due to \emph{measurement error}. Intuitively,  calibrating decisions to be fair for an error-prone proxy does not imply the decision is fair for the true variable of interest. Such effects can be substantial; for example, \cite{obermeyer_dissecting_2019} demonstrated a large differential in predicted risk scores between black and white patients with equal values on a new proxy measurement. However, these authors did not suggest a method for dealing with this problem. This issue cannot be ignored because fairness is generally conceptualized on a level more abstract than the proxy label \cite{jacobs2019measurement}; for example, it is reasonable to require that fairness in a healthcare need prediction system should extend to a person's true health status. 

This paper addresses the problem of prediction unfairness arising from measurement error. By considering the supervised learning problem at the level of a latent variable of interest, we reformulate the problem as one of adequate \emph{measurement modeling}. In effect, instead of requiring perfect measurement to achieve fairness, we propose that researchers developing a prediction model to be used for decision-making collect several independent, possibly error-prone, measures of the variable of interest (e.g. health). We then suggest to combine measurement models from the statistical literature with techniques from the literature on fair ML to assess and ameliorate the problem of unfair predictions in the face of measurement error.

Our contributions are as follows:
\begin{itemize}
    \item We illustrate that existing methods to examine unfairness in error-prone outcomes are insufficient;
    \item We suggest a framework, based on the existing measurement modeling literature, to investigate and ameliorate such issues;
    \item We perform an example analysis to demonstrate the suggested approach. In an existing healthcare application, this demonstrates that replacing one proxy with another does not lead to parity, while our approach does.
\end{itemize}

In Section \ref{sec:problem}, we provide a summary of basic concepts in fairness. In Section \ref{sec:prior} prior approaches with respect to fair inference are discussed. In Section \ref{sec:measurement}, the failure of these approaches is discussed when making use of proxies, and the proposed framework is introduced based on existing measurement models. In Section \ref{sec:results} the proposed framework is then applied to the exemplary data set provided by \cite{obermeyer_dissecting_2019}.

\section{Problem definition} 
\label{sec:problem}
We consider probabilistic classification and regression problems with a set of features $\mathbf{X}$ and true outcome $Y^*$. Among the features, there is a sensitive feature $S \in \mathbf{X}$ (e.g. race, gender), with respect to which discriminatory predictions are to be avoided. Furthermore, although the prediction problem is with respect to the true outcome $Y^*$ -- e.g. ``health'' or ``crime'' -- this outcome is not directly observed; instead, we have observed a set of error-prone proxy variables $\mathbf{Y}$. For example, in practice a proxy for ``health'', $Y \in \mathbf{Y}$, might be the costs of healthcare or the number of chronic conditions experienced by the patient, whereas, instead of ``crime'', the number of arrests might be measured. Following \cite{nabi_fair_2018}, we represent the goal of the regression or classification problem as a query on the (generative) joint distribution $p(Y^*, \mathbf{X})$, potentially after conditioning on a set of ``fixed'' covariates $\mathbf{C}$, i.e. the (discriminative) conditional joint $p(Y^*, \mathbf{X} \setminus \mathbf{C} \mid \mathbf{C})$. Typically, this query will be the point prediction $\hat{Y}^* := E(Y^*\mid\mathbf{X})$.

Following standard social-scientific measurement theory \cite{borsboom2006does}, the fact that $\mathbf{Y}$ is a measurement proxy for $Y^*$ is reflected by a \emph{causal model}, in the sense of \cite{pearl_causality_2013,spirtes2000causation}, in which $Y^* \rightarrow \mathbf{Y}$, i.e. the true outcome is a common cause of all available proxy variables. Because $Y^*$ is an unobserved latent variable, our causal model will be identifiable only through additional assumptions of conditional independence; we  discuss these assumptions later. The key point to note here is that, generally, $E(Y^*\mid\mathbf{X}) \neq E(Y \in \mathbf{Y}\mid\mathbf{X})$, i.e.  predictions using error-prone proxies as labels, $\hat{Y}$, will, of course, differ from the $\hat{Y}^*$ that would have been obtained had the true labels been available.

\section{Related work}
\label{sec:prior}
A large and growing literature on fairness of predictions for the error-free outcome $Y^*$ exists, with divergent and sometimes mutually exclusive definitions of the notion of algorithmic fairness. An excellent overview of this literature can be found in \cite{verma_fairness_2018}, which identified 20 separate definitions. Broadly, a distinction can be made between statistical metrics, distance-based measures, and causal reasoning \cite{verma_fairness_2018}.

Statistical metrics define fairness as the presence or absence of a (conditional) independence in the joint distribution $p(Y^*, \hat{Y}^*, S)$. For example, take  a classification problem in which the decision is taken as $d := I(\hat{Y}^* > \tau)$, where $I$ is the indicator function and $\tau$ is some threshold on the predicted score. \emph{Statistical parity} (``group fairness'') is then defined as $p(d=1\mid S=s) = p(d=1\mid S=s^\prime)$ for all $s\neq s^\prime$, i.e., the decision should not depend on the sensitive attribute, whereas \emph{predictive parity} is defined as $p(Y^*=1\mid d=1, S = s) = p(Y^*=1\mid d=1, S = s^\prime)$  for all $s\neq s^\prime$---i.e. the positive predictive value should not depend on the sensitive attribute. Further definitions include conditional statistical parity \cite{corbett2017algorithmic}, overall accuracy equality \cite{berk2018fairness}, and well calibration \cite{kleinberg2016inherent}.

Distance-based measures of fairness account for the non-sensitive predictors $\mathbf{X}\setminus S$, in addition to the observed and predicted outcomes and sensitive attribute. The well-known ``fairness through awareness'' framework \cite{dwork2012fairness} generalises several of the preceding notions, such as statistical parity, by defining fairness as ``similar decisions for similar people''. Consider a population of potential applicants $P$, and consider any randomised output from the prediction algorithm, $M(x \in P)$. Fairness is achieved whenever the distance among the decisions $M$ made for two people is at least as small as the distance between these people, i.e. when $D(M(x), M(y)) \leq d(x, y)$ for any $x, y \in P$. Here, $D$ and $d$ are arbitrary metrics on the distance between outputs and people, respectively. Careful choice of these metrics can yield some of the above definitions as special cases. Since the fairness condition can be trivially achieved, for example by always outputting a constant regardless of the input, the prediction model should be trained by minimising a loss function under the above constraint.

\begin{figure}[H]
\centering
\includegraphics[width=\linewidth]{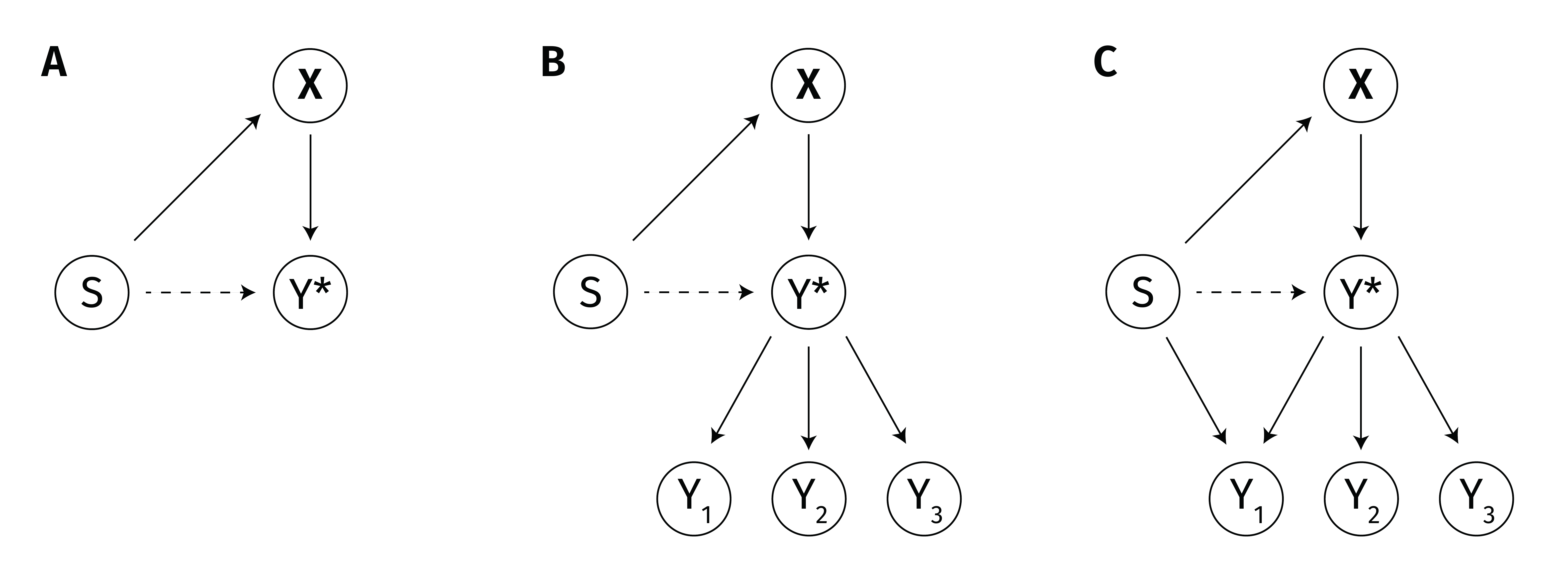}
\caption{Graphical representation of causal relations between the sensitive feature ($S$), the predictors (${\bf X}$), and the error-prone outcome ($\hat{Y}^*$) in the naive case (A), in the measurement error framework (B), and in the measurement error framework with differential item functioning on the $Y_1$ proxy (C). The dotted arrow indicates the discriminatory causal pathway (as in \cite{nabi_fair_2018}) which is blocked when performing fair inference, evaluating $E[Y^*\mid {\bf X}, S]$ to compute a risk score $\hat{Y}^*$.}
\label{fig:dags}
\end{figure}

Finally, in recent years, results from the causal modelling literature have been leveraged to define and achieve ``counterfactual'' fairness \cite{kusner_counterfactual_2017,nabi_fair_2018}. In these definitions one first considers a causal model involving $Y$, $X\setminus S$, and $S$ such as Panel A of Figure \ref{fig:dags}. This causal model then induces a counterfactual distribution $p_{do(s)}(\hat{Y}^*\mid X)$, i.e. the distribution we would observe if $S$ were \emph{set} to the value $s$ \cite{pearl_causality_2013}. \cite{kusner_counterfactual_2017} then defined counterfactual fairness as $p_{do(s)}(\hat{Y}^*\mid X) = p_{do(s^\prime)}(\hat{Y}^*\mid  X)$. Note that this definition looks superficially similar to the definition of statistical parity (group fairness), but is distinct because it refers to an individual. This definition has as a disadvantage that \emph{any} causal effect of the sensitive attribute on the prediction is deemed illegitimate. Based on the same framework, \cite{nabi_fair_2018} suggested a more general definition: some causal \emph{pathways} originating in $S$ are denoted discriminatory, while others are not. Fairness is then achieved by performing inference on a distribution $p^*(Y^*, \mathbf{X})$, in which the ``fair world'' distribution $p^*(Y^*, \mathbf{X})$ is close in a Kullback-Leibler sense to the original $p(Y^*, \mathbf{X})$, but all discriminatory pathways have been blocked (up to a tolerance) using standard causal inference techniques. Note that, if all causal pathways originating in $S$ are deemed discriminatory and the tolerance set to zero, the counterfactual fairness criterion by \cite{kusner_counterfactual_2017} will be satisfied.

\section{Proposed framework}
\label{sec:measurement}
\subsection{Fair inference in error-prone outcomes}
\label{sec:failure}
The existing methods from Section \ref{sec:prior} do not consider the target $Y^*$ to be error-prone. However, in practice, the target feature $Y \in {\bf Y}$ in the data set is not a perfect representation of the true underlying outcome $Y^*$. There can be several sources for this imperfect representation. For example, the true underlying outcome of interest may not be directly measurable at all (i.e., $Y^* \neq Y$ for any possible $Y$). In this case, the outcome of interest will only partially explain any feature used as its proxy. For example, in using healthcare costs $Y$ as a proxy for health $Y^*$, the observed value will in part be determined by other factors besides $Y^*$, such as the location of residence of the patient. Then, even if the outcome of interest were ``true healthcare costs'' -- thus in principle measurable -- the observed feature will in practice still not be an infallible proxy, because health records are never perfect observations and always contain some form of noise \cite{brakenhoff2018measurement}. Together, such sources of noise in the observation process are termed ``measurement error'', and any outcome $Y^*$ containing measurement error can be considered \emph{latent} \cite{borsboom2008latent} and modelled as such. 

Crucially, the presence of measurement error may result in unfair inferences for the error-prone outcome, even after applying the procedures presented in Section \ref{sec:prior} to account for unfairness. This is shown in a compelling example by \cite{obermeyer_dissecting_2019}, who concluded that commercial algorithms used by insurance companies for patient referral contain a fundamental racial bias. In the algorithm under consideration, healthcare costs $Y \in {\bf Y}$ are used as a proxy for health $Y^*$. \cite{obermeyer_dissecting_2019} illustrated that although there is no bias in healthcare costs, there is strong racial bias in other proxies of health such as whether patients have chronic conditions. Specifically, in order to be referred to a primary care physician, the true underlying health status $Y^*$ of black patients was worse than that of white patients.

\cite{obermeyer_dissecting_2019} concluded that fair inference requires selecting a better proxy for health as the outcome variable $Y^*$. Indeed, their analyses were possible precisely due to the availability of different proxies of health, such as the number of chronic conditions. However, we note that solving racial bias in a new proxy does not guarantee the absence of racial bias in other proxies indicating other aspects of health. Instead, here we suggest incorporating several proxies, or \emph{indicators} ${\bf Y}$ in a measurement model for the unobserved, error-prone outcome $Y^*$ \cite{kilbertus2017avoiding}. In the next section, we introduce the existing literature on measurement models and its approach to fair inference.

\subsection{Fair inference in measurement models}
\label{sec:measurementmodels}
When outcomes are thought to be error-prone, an existing literature suggests the use of measurement models \cite{fuller2009measurement,brakenhoff2018measurement}. At their core, measurement models describe the causal relationship between observed scores ${\bf Y}$ and error-prone unobserved ``true scores" $Y^*$ as $Y^* \to {\bf Y}$. A measurement model adequately represents the empirical conditions of measurement if conditional independence can be assumed \cite{blalock1968methodology}. More specifically, measurement models assume that $Y_1$ and $Y_2$ are conditionally independent given $Y^*$ (i.e., $p(Y_1, Y_2 \mid Y^*) = p(Y_1\mid Y^*) p(Y_2\mid Y^*)$). A plethora of variations of measurement models assuming conditional independence have been developed, such as latent class models \cite{mccutcheon1987latent}, item response models \cite{rasch1993probabilistic}, mixture models \cite{mclachlan1988mixture}, factor models \cite{lord2012applications},  structural equation models \cite{bollen_structural_1989}, and generalized latent variable models \cite{skrondal_generalized_2004}.

Measurement models are suggested here as a convenient way to account for a latent variable's relationship to sensitive features. The measurement error of a proxy variable (e.g. $Y_1$) is then assumed to differ over different groups of $S$. To account for group differences in proxy variables, a large body of literature is available where this issue is known under different labels. Generally, these approaches are applied within the structural equation modelling (SEM) framework \cite{joreskog1993testing}, as SEM explicitly separates the measurement model $(Y^* \to {\bf Y})$ from the structural model $({\bf X} \to Y^*)$. Approaches for investigating how features $S$ influence $Y^*$ are investigating item bias \cite{mellenbergh_item_1989}, Differential Item Functioning (DIF) \cite{holland_differential_1993} and measurement invariance \cite{schmitt2008measurement}. For an extensive overview of the different approaches and their benefits and drawbacks, we refer to \cite{flore2018stereotype,schmitt2008measurement,steenkamp1998assessing,vandenberg2000review}.

\subsection{Proposed method for fair inference on latent variables} 
\label{sec:framework}
% Here, we explain the SEM model we propose
We propose our framework for fair inference on outcomes which are measured only through error-prone proxies. To clarify the framework and make it more comparable to earlier work, we use the running example of health risk score prediction from \cite{obermeyer_dissecting_2019}. Their healthcare data set contains several clinical features ${\bf X}$ at time point $t-1$ (e.g., age, gender, care utilisation, biomarker values and comorbidities) which are used to predict healthcare cost $Y^*$ at time $t$. In addition, the patient's race is the sensitive feature $S$, coded as $S=b$ for black patients and $S=w$ for white patients. The relations between these features are shown in panel A of Figure \ref{fig:dags}. 

Based on ${\bf X}$, the expectation of a persons' healthcare cost is used as a risk score $\hat{Y}^* := E[Y^* \mid {\bf X}, S]$. The risk score is used to make a decision $D$ to refer a patient to their primary care physician to consider program enrolment. More specifically $d=1$ if $\hat{Y}^*$ is above the $55^{th}$ percentile. In this setting, attributes ${\bf X}$ can be legitimately controlled. However, conditional on ${\bf X}$ both groups in $S$ should have equal probability of being referred: $\text{P}(d=1 \mid {\bf X} = x, S = b) = \text{P}(d=1 \mid {\bf X} = x, S = w)$. As mentioned in Section \ref{sec:failure} and shown by \cite{obermeyer_dissecting_2019}, this procedure leads to bias in other proxies of $Y^*$, such as a patient's number of chronic conditions.

\begin{figure}
\centering
\includegraphics[width=0.9\linewidth]{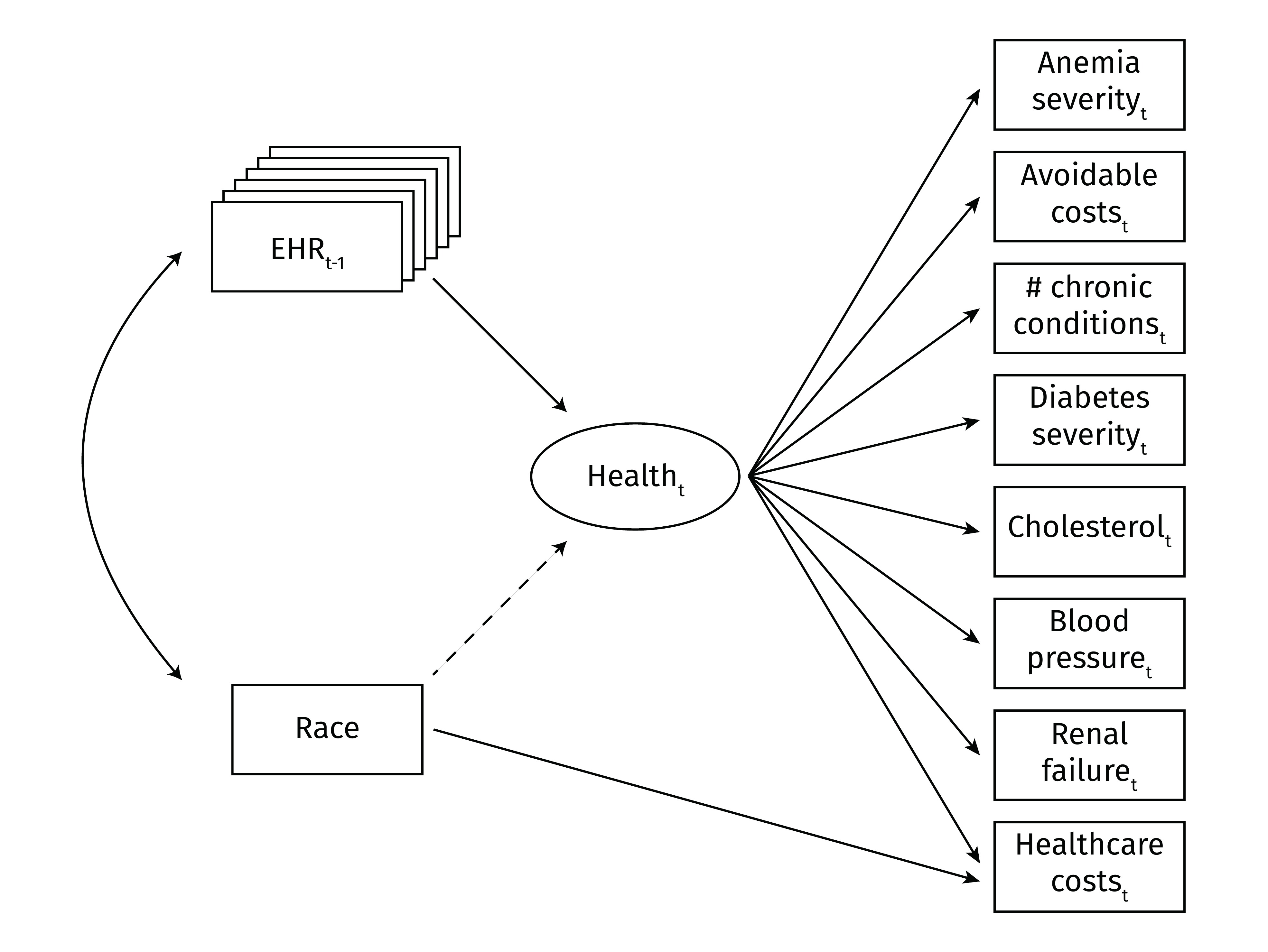}
\caption{Structural equation model for the proposed framework on the healthcare data set. For clarity, residual variances of the endogenous variables are not drawn in the diagram. For more information on the variables used in the model, see \cite{obermeyer_dissecting_2019}.}
\label{fig:SEM}
\end{figure}

Our proposed framework is a SEM implementation of the second and third panels of Figure \ref{fig:dags}. The general structure of the model is that of a Multiple Indicator, Multiple Causes (MIMIC) model: the outcome variable $Y^*$ (e.g., health) has multiple proxy indicators (e.g., chronic conditions, healthcare costs, hypertension), and the ${\bf X}$ features predict $Y^*$ directly (thus the proxies only indirectly). A graphical representation of the MIMIC SEM model is shown in Figure \ref{fig:SEM}. This implementation imposes additional assumptions on the general causal graphs, most notably linear relationships between the variables and the multivariate Gaussian residuals.

Fair inference on $Y^*$ can be performed in the following way: during estimation of the regression parameters (${\bf X} \to Y^*$), health is conditioned on race, but during prediction the path from Race to Health is blocked by setting $S=b$. Following the notation of \cite{nabi_fair_2018}, this yields a ``fair world'' distribution $p^*(Y^*, {\bf X})$. The expectation $\hat{Y}^* = E[Y^* \mid {\bf X}, S]$ is then computed from this distribution, meaning for two participants who differ only on $S$ but not on ${\bf X}$, the risk score $\hat{Y}^*$ will be exactly the same. Because the latent outcome $Y^*$ is modelled as a linear combination of the different proxies, the risk score is a reflection of the underlying health rather than only health cost.

\section{Experiments}
\label{sec:results}
In this section, we evaluate the proposed framework on an application of the procedures discussed in this paper. We first prepare the data set as provided by \cite{obermeyer_dissecting_2019} to create a basic risk score based on healthcare cost similar to the commercial risk score reported in their paper. Then, we illustrate our argument from Section \ref{sec:failure}: we perform fair inference on the proxy measure for health (healthcare cost) to show that this does not solve the issue of unfairness in other proxy measures. This is a reproduction of the results shown by \cite{obermeyer_dissecting_2019}. Next, we use the SEM framework from Section \ref{sec:framework} to show how including a formal measurement model for $Y^*$ -- as in panel B of Figure \ref{fig:dags} -- can largely solve the issue of unfairness in the proxies. Last, we show how existing differential item functioning (DIF) methods in the SEM framework  -- panel C of Figure \ref{fig:dags} -- can aid in interpreting the extent to which proxy measures contain unfairness. Fully reproducible R code for this section is available as supplementary material to this paper at the following DOI: \href{https://doi.org/10.5281/zenodo.3708150}{\texttt{10.5281/zenodo.3708150}}.

\subsection{Data preparation and feature selection} \label{sec:features}
% Include general description of data set first
Log-transformations are applied to highly skewed variables at time-point $t$, such as costs, to meet the assumption of normally distributed residuals in regression procedures. As an additional normalisation step, the predictors at time-point $t-1$ are re-scaled to homogenise their levels of variance. The data set is then split into a training and a test set. In this section, estimation is always done on the training set and inference is done on the test set.

To simplify our proposed framework for the purpose of this application, we select a subset of features at time-point $t-1$ for prediction of the target of interest at time point $t$, health. We want our procedure to be comparable to the commercial algorithm which produces the risk scores described in \cite{obermeyer_dissecting_2019}. If the features we select are the same features used by the commercial algorithm, then our procedure would yield very similar results upon generating a risk score. Unfortunately, the predicted risk scores used by \cite{obermeyer_dissecting_2019} cannot be replicated exactly using the provided data set. 

To select the subset of predictor features for further use in our procedure, we performed a LASSO regression  \cite{tibshirani2005sparsity} where all available features at time-point $t-1$ are used as predictor variables, and the provided algorithmic risk score at time-point $t$ is used as a target. Following the guidelines by \cite{hastie2009elements}, we used cross-validation to select the optimal $\lambda$ penalty value. This yields a set of non-zero predictors which predict the algorithmic risk score well.

% \begin{figure}
% \centering
% \includegraphics[width=0.7\linewidth]{01_feature_selection.png}
% \caption{Relationship between the commercial risk scores and the risk scores replicated using LASSO regression.}
% \label{fig:rep_risk}
% \end{figure}

%To illustrate how well the commercial risk scores are replicated, the relationship between the commercial risk scores and the risk scores replicated by means of our LASSO regression can be found in Figure \ref{fig:rep_risk}.
Spearmans rank correlation between the commercial and the replicated risk score is high $\rho = .82$, indicating that the commercial and replicated risk scores perform similarly in the rank-based cutoff applied in \cite{obermeyer_dissecting_2019}. The predictors selected in this model are used as predictors $\boldsymbol{X}$ in the structural equation models of the following sections.

\subsection{Fair inference on cost as a proxy of health} \label{sec:parity}

Pane A of Figure \ref{fig:dags} illustrates conditional statistical parity as defined by \cite{verma_fairness_2018}. Here, the outcome $Y^*$ is conditioned on sensitive feature $S$ when estimating the coefficients of the prediction model ($\boldsymbol{X} \to Y^*$), and during prediction all subjects are assumed to have the same level of $S$, e.g., $S = b$, such that $\text{P}(Y^*=y^* \mid {\bf X} = x, S = b) = \text{P}(Y^*=y^*\mid {\bf X} = x, S = w)$.

\begin{figure}
    \centering
    \includegraphics[trim={13cm 0 0 0},clip,width=\linewidth]{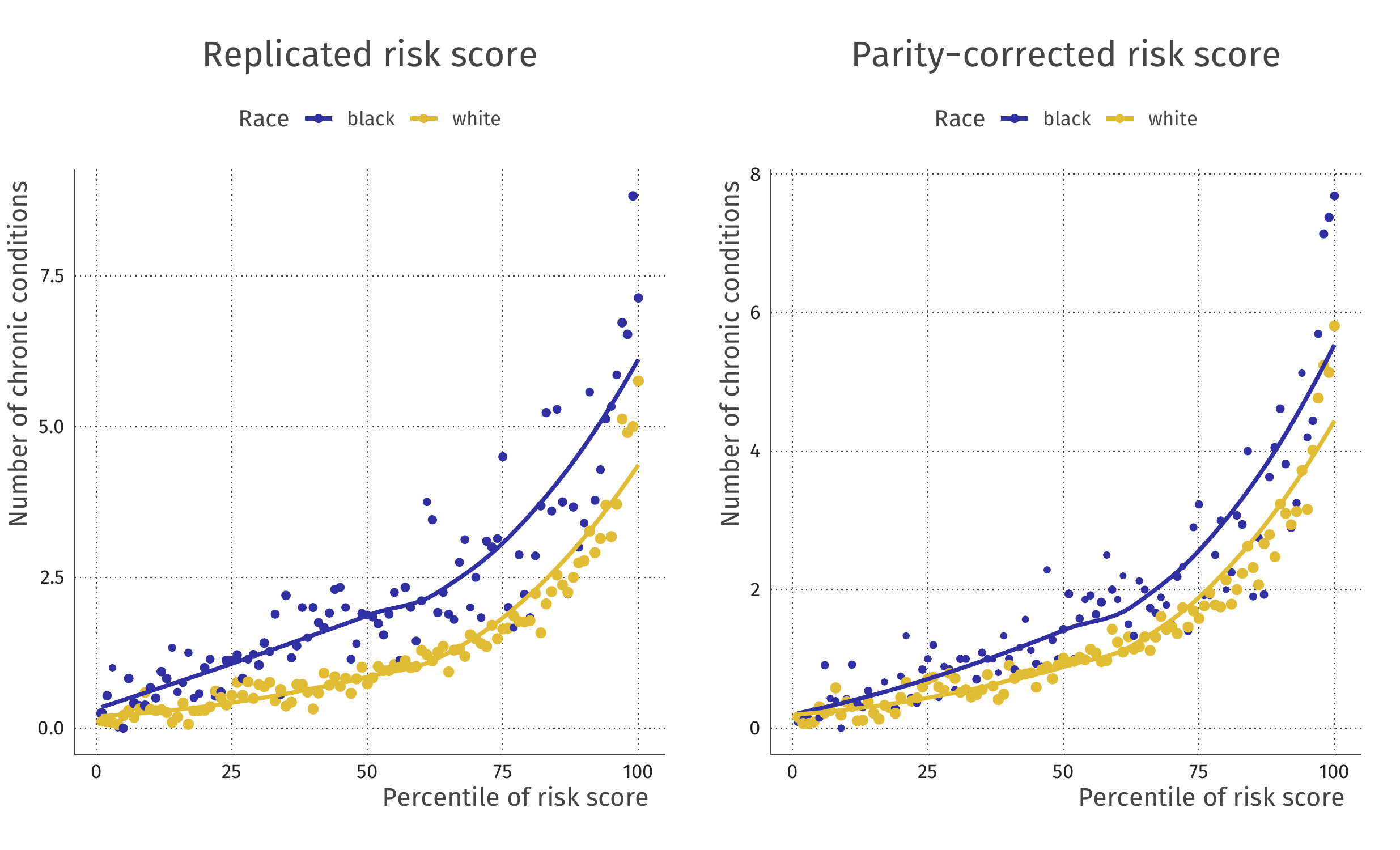}
    \caption{Effect of parity correction in one proxy of health (healthcare cost) on the race differences in another proxy of health (the number of chronic conditions). From the replicated risk score to the parity-corrected risk score, the cross-race difference becomes slightly smaller but does not disappear.}
    \label{fig:parity}
\end{figure}

Accounting for sensitive feature `Race' by conditioning the outcome `Replicated Risk score' on `Race' when estimating the model and being excluded during prediction reduces the extent of the problem. Figure \ref{fig:parity} illustrates that although the results improve compared to not including `Race' at all, conditional statistical parity is still not met. As a consequence, individuals belonging to $S=b$ will still have a lower health status when being selected for intervention.

\subsection{Fair inference on latent health} 
A cause for the fact that conditional statistical parity is not met when following Pane A of Figure \ref{fig:dags} can be that $\hat{Y}^*$ is a (bad) proxy. Instead of using one bad proxy, it is better to use multiple (bad) proxies as indicators of an unobserved latent variable measuring `true health'. How such a model can be specified is illustrated in Pane B of Figure \ref{fig:dags}. Similarly to \cite{verma_fairness_2018}, the sensitive feature is excluded during prediction.

\begin{figure*}
    \centering
    \includegraphics[width=\textwidth]{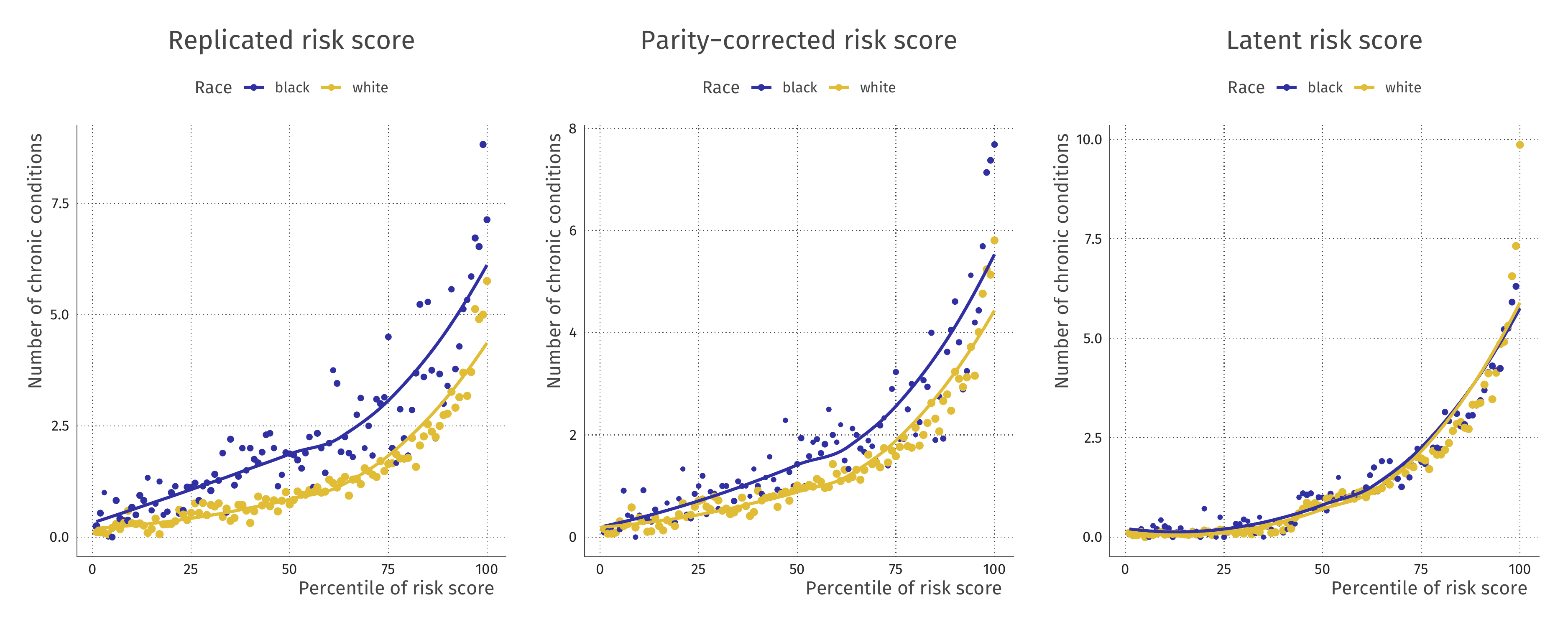}
    \caption{Effect of including a measurement model in constructing risk scores. The first panel shows the uncorrected risk score based on healthcare cost, the middle panel shows the same risk score but corrected for the sensitive feature, and the third panel shows the corrected risk score based on the latent health outcome using a measurement model.}
    \label{fig:measurement}
\end{figure*}

Figure \ref{fig:measurement} shows the effect of including a measurement model in constructing risk scores. This figure illustrates that using a measurement model with multiple imperfect measurements of health as indicators for `true health' substantially improves conditional statistical parity. The improvement has been more  compared to accounting for the sensitive feature. By using this measurement model, the problem that individuals belonging to $S=b$ had a lower health status when being selected for intervention is minimised.

\subsection{Investigating unfairness in proxies} 
When using a measurement model with multiple imperfect measurements of health as indicators of `true health', differences in measurement error over the different groups of the sensitive feature can still be present. Panel C of Figure \ref{fig:dags} illustrates how differences over the sensitive feature groups in the error prone indicator variables can be incorporated directly when estimating `true health'. For example, differences in measurement error of healthcare cost can be present for the different groups of Race. 

Including a DIF parameter $\delta$ on the healthcare cost variable yields a model which fits significantly better on the test set than the model without the DIF parameter ($\chi^2(1) = 50$, $p < 0.001$). The value of the DIF parameter on cost is estimated as $\delta = 0.198$ (95\% CI = $[0.172, 0.225]$). This means that for the same level of health, the log-healthcare costs of the white race class in this data set is estimated to be 0.198 higher. This means that the cost of healthcare for white patients is $(e^{0.198} - 1) \cdot 100\% = 21.9\%$ higher than that for black patients, \emph{given an equal level of health} as measured by the measurement model (95\% CI = $[18.7, 25.2]$).

Applying the same procedure to the other indicators leads to estimates of DIF for those indicators. The results are shown in Table \ref{tab:dif}. This table shows that some proxies have stronger DIF than others, meaning some proxies are more unfair than other proxies. Notable, the avoidable healthcare cost and the renal failure items have low levels of DIF for Race, whereas the healthcare cost and the number of active chronic conditions have strong DIF.

% latex table generated in R 3.6.2 by xtable 1.8-4 package
% Tue Mar 10 11:49:41 2020
\begin{table}
\centering
\caption{Estimated differential item functioning parameters for each indicator (proxy) of health. $\delta$ parameters should be interpreted as the mean deviation of the black patients compared to the white patients given health. \break}
\label{tab:dif}
\begin{tabular}{lccc}
  \hline
  Indicator & $\delta$ & 2.5\% & 97.5\% \\ 
  \hline
  No. active chronic conditions & 0.453 & 0.364 & 0.541 \\ 
  Mean blood pressure & -0.262 & -0.320 & -0.204 \\ 
  Diabetes severity (HbA1c) & -0.343 & -0.391 & -0.296 \\ 
  Anemia severity (hematocrit) & 0.250 & 0.231 & 0.268 \\ 
  Renal failure (creatinine) & -0.019 & -0.025 & -0.014 \\ 
  Cholesterol (mean LDL) & -0.235 & -0.317 & -0.153 \\ 
  Healthcare cost (log) & 0.198 & 0.172 & 0.225 \\ 
  Avoidable healthcare cost (log) & -0.052 & -0.096 & -0.008 \\ 
  \hline
\end{tabular}
\end{table}

\section{Conclusion}

In this paper, we have argued that when measurement error is at play, performing fair inference on a proxy measure of the outcome is insufficient to achieve a fair inference on the true outcome. This manifests itself, as shown in \cite{obermeyer_dissecting_2019}, as unfairness in other proxy measures of the outcome of interest. Alternatively, in this study we proposed to make use of existing measurement models containing multiple error-prone proxies for the outcome of interest. In addition, fair inference can be accounted for in each of these proxies simultaneously if needed by allowing for measurement error in proxies to differ over groups of a sensitive feature. We provided a framework to perform these estimations and applied this framework to the exemplary data set provided by \cite{obermeyer_dissecting_2019}. Here, it was concluded that fair inference was accounted for when multiple proxies were used in a measurement model instead of a single proxy. Additionally accounting for differences in measurement error over race groups was not needed to further improve fairness in predicted risk scores, although substantive group differences were found for some proxies.

\bibliographystyle{plain}
\bibliography{fair}
\end{document}